# Knowledge Pyramid: A Novel Hierarchical Reasoning Structure for Generalized Knowledge Augmentation and Inference

Qinghua Huang, Yongzhen Wang

*Abstract*—Knowledge graph (KG) based reasoning has been regarded as an effective means for the analysis of semantic networks and is of great usefulness in areas of information retrieval, recommendation, decision-making, and man-machine interaction. It is widely used in recommendation, decision-making, question-answering, search, and other fields. However, previous studies mainly used low-level knowledge in the KG for reasoning, which may result in insufficient generalization and poor robustness of reasoning. To this end, this paper proposes a new inference approach using a novel knowledge augmentation strategy to improve the generalization capability of KG. This framework extracts high-level pyramidal knowledge from low-level knowledge and applies it to reasoning in a multi-level hierarchical KG, called knowledge pyramid in this paper. We tested some medical data sets using the proposed approach, and the experimental results show that the proposed knowledge pyramid has improved the knowledge inference performance with better generalization. Especially, when there are fewer training samples, the inference accuracy can be significantly improved.

## I. INTRODUCTION

A knowledge graph (KG) is a structured representation of facts, entities, relationships, and semantic descriptions [1]. KG completion is an important research direction in KGs. KG reasoning belongs to the category of knowledge graph completion and has been widely used in various important fields, with various forms of reasoning. Using the data form of KG for knowledge reasoning has the advantage of tolerating knowledge loss [2]. The current work on reasoning in the form of KG is mainly divided into three categories: rule-based reasoning methods, representation learning-based methods, and neural network and reinforcement learning-based reasoning methods. The rule-based reasoning method mainly integrates logical rules into the embedding and uses simple rules or statistical features for reasoning. The relational path rule-based reasoning method mainly uses the path information in the graph structure to search for complex relational path information and implement inference. The method based on representation learning is to embed high-dimensional entity and relation attributes into the same low-dimensional space, score candidate entities through the scoring function, and realize KG reasoning through scoring. The reasoning method based on neural network mainly transforms the feature distribution of the input data from the original space to another feature space through nonlinear transformation and automatically learns the feature representation for reasoning. Another method based on reinforcement learning is mainly converting. The path discovery problem between entity pairs is formalized as a sequential decision problem. The agent finds a relational step through the interaction between the knowledge graph environment and expands the reasoning path to realize multi-hop reasoning. [3] The above methods all have their own shortcomings. For example, both representation learning and neural network-based methods lack reasoning interpretability, and rule-based methods are not effective in reasoning. [1] The above methods are all improved at the model level, and there is no improvement in the direction of original data augmentation, indicating that there is still space for further improvement in the accuracy of inference.

In this study, we propose a new KG-based data augmentation model (KGDA). The whole work is divided into three parts: data augmentation, data fusion, and augmented model performance verification. Data augmentation refers to a layer of data augmentation based on the original structure of the KG data model, forming a hierarchical KG structure called the Knowledge Pyramid (KP) in this paper. Data fusion is to perform a set operation on the augmented data to realize the fusion of the original data and the augmented data. The performance research of the augmented model mainly selects a certain inference model and compares the performance of the augmented model on the inference effect before and after augmentation.

For data augmentation of the KG, we use the idea of biclustering to realize it. The relationship between different features is mined from the original data through the biclustering algorithm, and new high-dimensional features are clustered. Each clustered bicluster vector is a new data feature compared to the original data. For data fusion, we first convert the augmented data into the form of a KG and then use a set operation to combine the augmented knowledge graph data and the original graph data to realize the fusion of knowledge

This work was supported in part by the National Key Research and Development Program under Grant 2018AAA0102100

Q. Huang and Y. Wang are with School of Artificial Intelligence, OPtics and ElectroNics (iOPEN), Northwestern Polytechnical University, Xi'an 710072, Shannxi, China. E-mail: qhhuang@nwpu.edu.cn, 1643030440@qq.com.

graph data before and after augmentation. For the performance verification of the augmented model, we use breast ultrasound data for testing. As for the inference model, we employ the TuckER model to detect the augmented data model and demonstrate the superiority of the data augmentation model by comparing the data before and after augmentation and setting different proportions of training data. In our comparative experiments, our proposed data augmentation model outperforms the inference without augmentation model. In the comparative experiments of different proportions of training data, the data augmentation model shows significant improvement in inference performance, particularly in the case of a few-sample training set.

II. BACKGROUND AND RELATED WORK

There are currently three main methods for reasoning in the form of knowledge graphs: rule-based reasoning, representation learning-based, and neural network and reinforcement learning-based reasoning.

*A. Rule-based reasoning*

The rule-based knowledge reasoning model is based on the fundamental concept of using simple rules or statistical features for reasoning. FOIL[4] explores all relations in the knowledge graph to derive a set of Horn clauses for each relation. These Horn clauses represent feature patterns that predict the existence of corresponding relations. Finally, a relational discriminant model is obtained using machine learning techniques. NELL[5] learns probabilistic rules, which are then instantiated after manual screening. This process enables the inference of new relation instances from other learned relation instances. PRA[6] incorporates path rules into knowledge reasoning and predicts missing edges in the graph by identifying edge-type sequences of linked nodes as features in a logistic regression model. CPRA[7], an extension of PRA [6], is a multi-task learning framework comprising two modules: relational clustering and relational coupling. The former automatically discovers highly correlated relationships, while the latter couples the learning of these relationships. However, this method suffers from poor generalization due to the manually defined rules.

*B. Representation Learning-Based Methods*

The representation learning methods primarily rely on embedding techniques, which map entities, relations, and attributes in a semantic network to continuous vector spaces using models to obtain distributed representations. RESCAL [8] decomposes high-dimensional and multi-relational data into third-order tensors, reducing the data dimension while preserving the essential characteristics of the original data. It successfully infers missing triples in the original graph through factorization. TransE[9] utilizes the Euclidean distance as a rigid metric for reasoning but faces limitations when dealing with complex relations. To address these limitations, TransAH[10] replaces the weighted Euclidean distance by incorporating a diagonal weight matrix and introduces a relation-oriented hyperplane. Experimental results on large-scale knowledge graphs demonstrate that TransAH is suitable for reasoning tasks. Inference methods based on GCN[11] typically exhibit linear scalability with the number of graph edges. They learn hidden layer representations that encode local graph structure and node features to facilitate reasoning. SME[12] uses vectors to represent entities and relationships from the perspective of semantic matching. It models the association between entities and relationships as a semantic matching energy function and defines both linear and bilinear forms for the semantic matching energy function. However, representation learning methods may lack interpretability since they do not provide intuitive reasoning processes.

*C. Methods Based on Neural Networks and Reinforcement Learning*

Neural network-based methods leverage the power of artificial neural networks to transform the feature distribution of input data from the original space to another feature space through nonlinear transformations. They automatically learn feature representations for inference tasks. MLP [13] employs fully connected layers to encode entities and relations and utilizes a second layer with nonlinear activation functions to score triples. The SLM [14] model reasons about the relationship between two entities by leveraging the nonlinear implicit connections between entity vectors in standard single-layer neural networks. The NTN [15] model replaces the linear neural network layer with a bilinear tensor layer, directly linking two entity vectors across multiple dimensions. It initializes entity representations by averaging word vectors, leading to improved performance. The ProjE [16] model achieves a smaller parameter size by making simple structural changes to the shared variable neural network model. The NAM [17] model represents all symbolic events in a low-dimensional vector space for probabilistic reasoning in artificial intelligence. Reinforcement learning-based methods tackle knowledge reasoning by formulating the problem of path discovery between entity pairs as a sequential decision problem. DeepPath [18] encodes states in a continuous space using translational embedding, employs the relation space as the action space, and introduces a novel reward function to enhance accuracy, path diversity, and path efficiency. Multi-Hop [19], instead of using a binary reward function, proposes a soft reward mechanism. It also utilizes action dropout during the training process to mask some outgoing edges and improve path search efficiency. M-Walk [20] incorporates an RNN controller to capture historical trajectories and employs Monte Carlo Tree Search (MCTS) for efficient path generation. However, one drawback of neural network-based methods is their lack of interpretability, as they do not provide an intuitive reasoning process.

*D. Knowledge Extraction*

Knowledge extraction encompasses entity extraction and relationship extraction. Entity extraction aims to identify and locate entity information elements within text, followed by classifying the identified entities into predefined categories.

Deep learning methods, such as neural network structures, are often employed for entity recognition. These neural networks serve as encoders, generating new vector representations for each word based on initial input and contextual information. The CRF model[21] is then utilized to produce annotation results for each word, effectively performing entity extraction. Relationship extraction involves extracting semantic relationships between two or more entities from text. Joint extraction methods[22] combine entity extraction and relationship extraction. By inputting a sentence into a joint model of entity recognition and relationship extraction, a related entity triplet can be directly obtained. In the context of relationship extraction, Huang[23] employed the biclustering algorithm to mine data consistency, providing further insight into the effectiveness of the biclustering algorithm.

It can be seen from the previous research work that the existing reasoning methods have problems such as poor generalizability, insufficient generalization, and lack of interpretability. The biclustering method can extract high-order semantic features, reduce data noise, improve the level of knowledge representation, and realize knowledge augmentation and more generalized knowledge reasoning.

## III. METHODS

In this study, we present a novel framework called the KG-based data augmentation model (KGDA). The KGDA framework leverages the original knowledge graph to cluster the data and applies the biclustering algorithm to extract recurring joint representations as augmented data features. Each extracted high-dimensional vector in the framework represents a new feature attribute. Additionally, we calculate the Euclidean distance between each high-dimensional vector and the original data, serving as the eigenvalue for each expanded feature attribute. To evaluate the performance of the augmented data model, we fuse the augmented data with the original data using set operations. Fig.2 illustrates the KG-based data augmentation model proposed in this study. Furthermore, the augmented data can be abstracted into a Knowledge Pyramid (KP), providing a hierarchical representation of the enriched knowledge.

### A. Knowledge Pyramid Augmentation

Due to the insufficient data information of low-level knowledge, it may lead to insufficient generalization and poor robustness of inference. Therefore, we have designed a novel knowledge augmentation strategy to improve the generalization ability of inference. This method utilizes biclustering to explore the recurrent joint feature representations in the dataset and extract high-level features from lower order features. The main idea is to generate candidate seed points through hierarchical clustering, and then use heuristic search biclustering to obtain higher-order features.

For a given data set, there may be a problem that the range is too large, and in order to facilitate the learning of subsequent training parameters, it is necessary to normalize the data first, and then proceed to the next stage of the process. Since the original data is in the form of a two-bit matrix, we choose the normalization method of the maximum and minimum values. For the n-th feature K(m,n) of the m-th sample in the overall data, the normalized result value K' (m ,n ) can be expressed as:

$$K^{'}(m,n) = \frac{K(m,n) - K_{min}(n)}{K_{max}(n) - K_{min}(n)} \qquad (1)$$

For a certain column-equivalent biclustering, the values corresponding to each column of data in the cluster are equal. If only column values are clustered, the clustering result will be many one-dimensional clusters. Obviously, some biclusters may contain these one-dimensional clusters. Therefore, horizontal search expansion can be carried out through these one-dimensional clusters, and it can be extended to two-dimensional double-clustering. On this basis, the speed of searching for clusters can be greatly accelerated and the time complexity of the entire algorithm can be reduced. The idea of hierarchical clustering [24] is as follows: Hierarchical clustering adopts a bottom-up search and merging strategy, and continuously merges clusters with high similarity until all clusters are merged into one cluster, or the similarity between the clusters exceeds the set value and cannot be merged.

Similar to traditional clustering algorithms, biclustering algorithms aim to partition data into different subsets based on data similarity. However, the key distinction lies in the fact that traditional clustering algorithms cluster samples in the entire feature space, disregarding the local consistency of samples within specific features. As a result, the output of traditional clustering is typically a subset of samples. In contrast, biclustering algorithms simultaneously cluster both the feature and sample dimensions, emphasizing the local consistency of the data. The resulting clusters consist of subsets of both features and samples. Figure 1 illustrates the process of generating biclusters through hierarchical clustering. Firstly, bicluster seed points are generated, and the columns of these seed points are expanded to form a submatrix. Then, a greedy strategy is employed to iteratively remove rows and columns from the submatrix until the MSRS (Mean Squared Residue) value of the submatrix falls below a specified threshold. Finally, the column means of the identified biclusters are computed to obtain new high-order entity features.

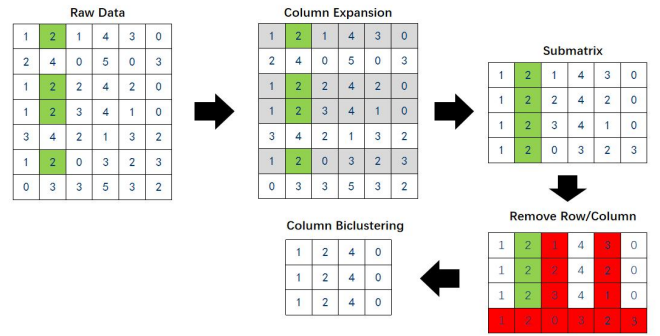

Fig.1. The process of generating biclusters

After performing hierarchical clustering, a heuristic search is performed starting from each one-dimensional cluster. The mean square residual score is used to evaluate the quality of the searched biclusters when searching. The definition of MSR is：

$$MSR = \frac{1}{|R||C|}\sum_{i\in R, j\in C}(x_{ij} - x_{Ri} - x_{iC} + x_{RC})^2 \quad (2)$$

$$x_{ic} = \frac{1}{|C|}\sum_{j\in C} x_{ij}, x_{Rj} = \frac{1}{|R|}\sum_{i\in R} x_{ij}, x_{RC} = \frac{1}{|R||C|}\sum_{i\in R, j\in C} x_{ij}$$

where $x_{ij}$ represents the value of the i-th row and j-th column of the biclustering matrix. where R and C represent the number of rows and columns of the biclustering matrix. The flow chart of the entire biclustering is as follows: First, expand the one-dimensional hierarchical clustering to all columns to obtain a result sub-matrix, and then calculate the MSR for the expanded sub-matrix. If the MSR exceeds the preset threshold, the greedy algorithm is used. Delete the row or column that can make the MSR score of the sub-matrix the smallest, repeat the above operation continuously to make the MSR less than the preset threshold, and finally perform a deduplication operation on all the found bicluster sets to remove the duplicated biclusters.

For the high-dimensional vectors obtained in the previous section, we calculate the Euler distance between each high-dimensional vector and the original data as the eigenvalues of each newly expanded feature attribute. The formula is as follows:

$$T_i = [v_{(1)}, v_{(2)}, v_{(3)}, \ldots v_{(j)}], \quad j \in S_i \quad (3)$$

$$F_i = \sqrt[2]{\sum_{i=1}^{n}(M_{(i,v(j))} - T_{(i,v(j))})^2}$$

where $v_{(i)}$ represents the i-th feature value contained in the dataset. $S_i$ represents the index of the i-th extracted high-dimensional feature in the dataset. $F_i$ represents the i-th new feature extracted. $M_{((i,v(j)))}$ represents the $v_{(j)}$ data feature of the i-th sample in the original data matrix.

*B. Data Fusion*

Since augmented data differs from the original data, it is necessary to combine the augmented data with the original data in order to evaluate the performance of the augmented model. This fusion process can only be achieved by converting diverse data forms into a unified format. Therefore, this paper proposes a method to transform both the original dataset and the augmented dataset into knowledge graph triples, enabling the merging of triples with the same format for data fusion. The procedure for fusing dimensional feature data is as follows:

$$G \in (E, R, S)$$
$$E \in \{e \mid e_o \cup e_a\}, R \in \{r \mid r_o \cup r_a\} \quad (4)$$
$$S \in S_o \cup S_a, S_o \in (e_o, r_o, e_o), S_a \in (e_a, r_a, e_a)$$

$S_o$ stands for the knowledge graph triple form of the original data, $S_a$ stands for the high-dimensional feature knowledge graph triple form extracted by the biclustering algorithm and ∪ stands for the union operation of the set operation. E, R, S respectively represent the fused entity, the fused relationship, and the fused triplet. After the set operation, the fused data are all in the form of knowledge graph triples, and each triple exists uniquely in the fused dataset, and the fused knowledge graph triples are in the form of: (patient ID, relationship, the relationship attribute value). Data fusion Process is shown in Fig.4.

*C. Inference Model*

To enhance the evaluation of our augmented data model, we drew inspiration from TuckER decomposition and opted to employ the TuckER decomposition model for verification. Initially, we employed the approach outlined in KTEAED [25] to convert the knowledge graph into a knowledge tensor format. This involved utilizing distinct layers to represent different relationships, resulting in hierarchical tensors. The TuckER decomposition[26] technique enables the decomposition of a tensor into a low-rank core tensor with a relatively smaller dimension and a series of matrices. The matrices represent embedding vectors of entities and relations, while the low-rank core tensor serves as a shared matrix for the entire set of entities and relationships. For a given three-dimensional vector $X \in R^{D1 \times D2 \times D3}$, TuckER decomposition can decompose the input vector into a low-rank core tensor $G \in R^{R1 \times R2 \times R3}$, and three-factor matrices $A^{D1 \times R1}$, $B^{D2 \times R2}$, $C^{D3 \times R3}$ [27]. Among them, D1, D2, and D3 are the three dimensions of the original data tensor, respectively, and R1, R2, and R3 are the three dimensions of the core tensor. Here, R1, R2, and R3 are much smaller than D1, D2, and D3. The formula is expressed as follows:

$$X \approx Z \times_1 A \times_2 B \times_3 C \quad (5)$$

Through TuckER decomposition, it can be found that the first and third matrix factors are used as subject embedding and object embedding respectively, and satisfy $A=C \in R^{m_e \times k_e}$, while the second matrix factor is used as relation embedding, and satisfy $B \in R^{m_r \times k_r}$, where $m_e$ and $m_r$ are the number of entities and the number of relationships, respectively. The dimensions are $k_e$ and $k_r$, respectively. Then the entire inference model can be further expressed as follows:

$$\phi(e_s, r, e_o) = W \times_1 e_s \times_2 w_r \times_3 e_o$$
$$P(e_s, r, e_o) = \sigma(\phi(e_s, r, e_o)) \quad (6)$$

In the above formula, W represents the shared low-rank matrix, $e_s$ represents the subject entity embedding, $e_o$ represents the object entity embedding, and $w_r$ represents the relationship embedding. The model structure is shown in Fig.5.The loss function of the entire model uses cross-entropy loss, where $y_{(i)}$ represents the true value of the i-th sample, $p_{(i)}$ represents the predicted probability of the i-th value, and $n_e$ is the number of entities, specifically expressed as follows:

$$L = -\frac{1}{n_e}\sum_{i=1}^{n_e}(y^{(i)}\log(p^{(i)}) + (1-y^{(i)})\log(1-(p^{(i)})) \quad (7)$$

Fig.5 shows the structure of the TuckER model. For the entire TuckER decomposition model, a Sigmoid function is added to the output of the entire inference model as the output of the entire inference model. Since the overall output value of the Sigmoid function is between 0 and 1, The output of the final model is the probability of inferring each entity.

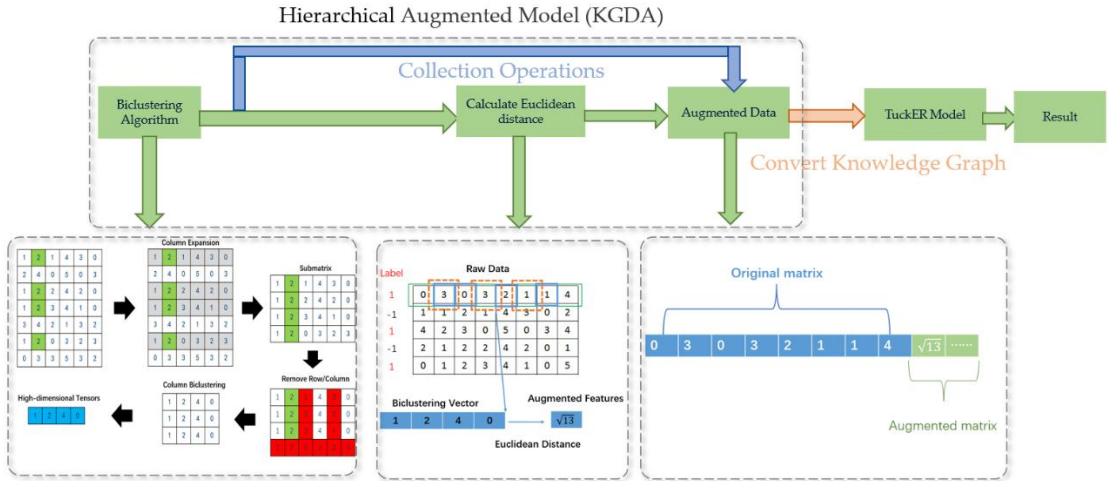

Fig.2. KGDA Model Structure. This figure shows the flow of the entire augmented model, which shows the biclustered process in detail, and the process of calculating the Euler distance augmented data.

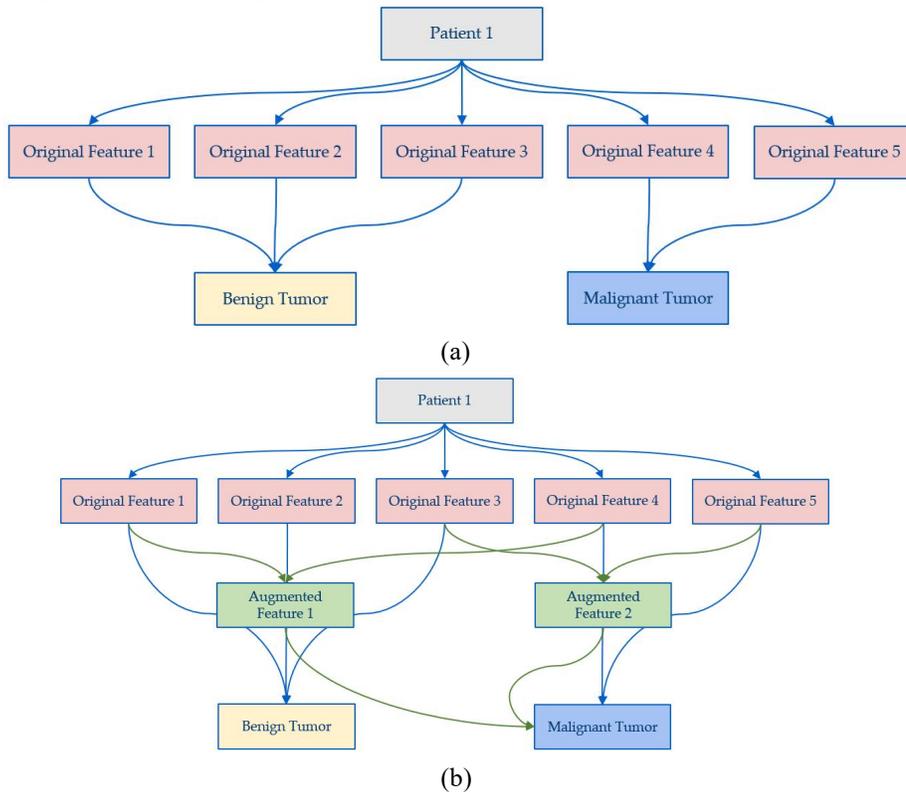

Fig.3. The difference between the original data model (a) and the augmented data model (b). The difference between the original data model and the augmented data models. It can be seen from the figure that the augmented high-dimensional features will help model inference.

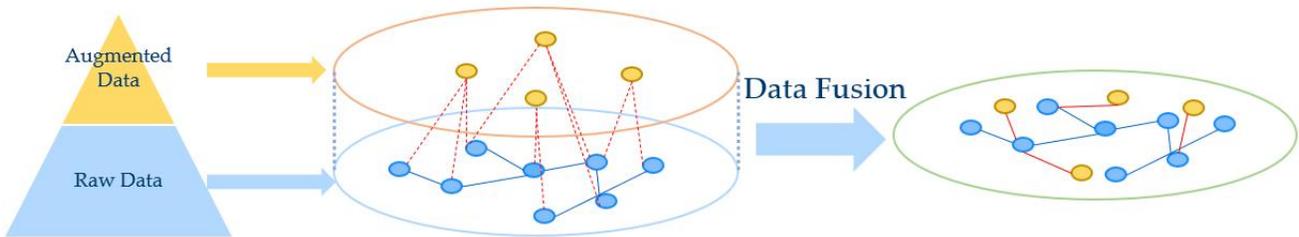

Fig.4. Diagram of the proposed knowledge pyramid structure for knowledge augmentation.

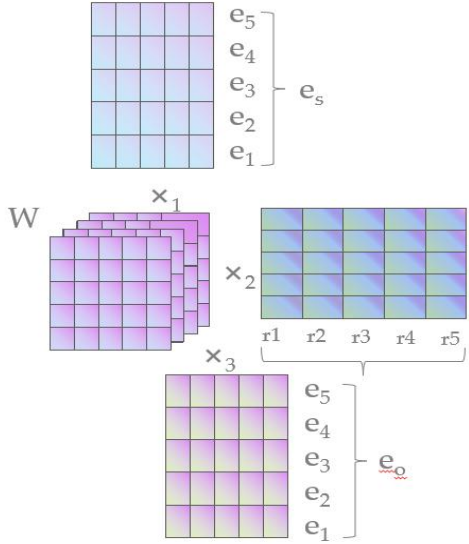

Fig.5. Diagram of the Tucker decomposition Model.

## IV. ANALYSIS OF THE MODEL PRINCIPLE

To evaluate the augmented model's performance, we conducted a series of comparative experiments on the breast ultrasound dataset, as illustrated in Fig.3. In contrast to the TuckER model, which directly utilizes data features for scoring low-level features to determine the benign or malignant nature of breast tumors, our approach incorporates the biclustering algorithm. This algorithm facilitates the extraction of joint representation features between low-level features, as depicted in Fig.3(a). Inferring the benign or malignant status of a breast tumor based on a single feature might lead to errors since some features exhibit combined characteristics. For instance, in Fig.3(a), the original data model mistakenly infers the tumor as benign due to the greater probability of benign tumors despite the last two features pointing towards malignancy, resulting in an inference opposite to the actual outcome. In contrast, the augmented model with bicluster augmentation extracts high-dimensional features by mining the joint representation features, as shown in Fig.3(b). The TuckER inference model utilizes these extracted high-dimensional features to infer the benign or malignant status of the current tumor. By incorporating the augmented high-dimensional features, which all indicate malignancy, the final inference correctly identifies the tumor as malignant, matching the actual result. Therefore, the augmentation of knowledge graph data using the biclustering model effectively uncovers deep-level feature combinations, leading to improved accuracy in model inference.

In mathematics, assuming that the clustered data distribution $X \sim N(\mu, \sigma^2)$ and the added noise data $Y \sim N(\nu, \tau^2)$ are independent of each other and obey the Gaussian distribution, then the original data set can be expressed as $Z = X + Y$. Then the probability function of Z can be expressed as follows:

$$F_z(z) = P(Z \leq z) = P(X + Y \leq z) = P(X \leq z - Y)$$

$$= \int_{-\infty}^{+\infty} f_Y(y) \int_{-\infty}^{z-y} f_X(x)dxdy = \int_{-\infty}^{+\infty} f_Y F_X(z-y)dxdy \quad (8)$$

The probability density function of Z can be expressed as:

$$f_Z(z) = F'_Z(z) = \int_{-\infty}^{+\infty} f_Y(y) f_X(z-y) dxdy \quad (9)$$

The probability density function of the Gaussian distribution can be expressed as:

$$f_X(x) = \frac{1}{\sqrt{2\pi}\sigma} e^{-\frac{1}{2\sigma^2}(x-\mu)^2} \quad (10)$$

Bringing the probability density function of the Gaussian distribution into the probability density function of Z further simplifies the calculation.

$$f_Z(z) = \int_{-\infty}^{+\infty} \frac{1}{\sqrt{2\pi}\tau} e^{-\frac{1}{2\tau^2}(y-v)^2} \frac{1}{\sqrt{2\pi}\sigma} e^{-\frac{1}{2\sigma^2}(x-\mu)^2} dy$$

$$= \frac{1}{2\pi\tau\sigma} \int_{-\infty}^{+\infty} e^{-\frac{1}{2}\left[\frac{1}{\tau^2}y^2 - 2\frac{1}{\tau^2}yv + \frac{1}{\tau^2}v^2 + \frac{1}{\sigma^2}y^2 + 2\frac{1}{\sigma^2}(\mu-z)y + \frac{1}{\sigma^2}(\mu-z)^2\right]} dy$$

$$= \frac{1}{2\pi\tau\sigma} \int_{-\infty}^{+\infty} e^{-\frac{1}{2}\left[(\frac{1}{\tau^2}+\frac{1}{\sigma^2})y^2 + (\frac{2}{\sigma^2}(\mu-z) - \frac{2}{\tau^2}v)y + \frac{1}{\tau^2}v^2 + \frac{1}{\sigma^2}(\mu-z)^2\right]} dy$$

$$= \frac{1}{2\pi\tau\sigma} \int_{-\infty}^{+\infty} e^{-\frac{1}{2}\frac{\sigma^2+\tau^2}{\sigma^2\tau^2}\left[y^2 + \frac{2\tau^2(\mu-z)-2\sigma^2 v}{\sigma^2+\tau^2}y + (\frac{\tau^2(\mu-z)-\sigma^2 v}{\sigma^2+\tau^2})^2 + \frac{\sigma^2\tau^2(\mu-z-\tau)^2}{(\sigma^2+\tau^2)}\right]} dy$$

$$= \frac{1}{\sqrt{2\pi}\sqrt{\sigma^2+\tau^2}} e^{\frac{(u-z+\tau)^2}{\sigma^2\tau^2}} \int_{-\infty}^{+\infty} \frac{1}{\sqrt{2\pi}\sqrt{\frac{\sigma^2\tau^2}{\sigma^2+\tau^2}}} e^{-\frac{1}{2}\frac{\sigma^2+\tau^2}{\sigma^2\tau^2}\left[y + \frac{\tau^2(\mu-z)-\sigma^2 v}{\sigma^2+\tau^2}\right]^2} dy$$

Through the above calculation, we can see that the integral part of the right side of the simplified equation satisfies the Gaussian distribution and its integral value is equal to one, that is:

$$y \sim N\left(-\frac{\tau^2(\mu-z)-\sigma^2 v}{\sigma^2+\tau^2}, \frac{\sigma^2\tau^2}{\sigma^2+\tau^2}\right) \quad (11)$$

So the probability density function of Z can be simplified as:

$$f_Z(z) = \frac{1}{\sqrt{2\pi}\sqrt{\sigma^2+\tau^2}} e^{-\frac{1}{2}\frac{(z-\tau-\mu)^2}{\sigma^2+\tau^2}} \quad (12)$$

From the above derivation, the solution can be drawn: the data of the original data Z set conforms to the Gaussian distribution, and the mean and variance are as follows:

$$Z \sim N(\tau + \mu, \sigma^2 + \tau^2) \quad (13)$$

Since the variance $\tau$ of the noise $Y \sim N(\nu, \tau^2)$ data is greater than 0, the variance of the data distribution $X \sim N(\mu, \sigma^2)$ after biclustering is smaller than the variance of the original data $Z \sim N(\mu + \nu, \sigma^2 + \tau^2)$, which is

$$\sigma < \sqrt{\sigma^2 + \tau^2} \quad (14)$$

It can be seen from the change in the variance of the data before and after the augmentation that the variance of the augmented data is smaller than that of the original data, and the data distribution is more concentrated, so the accuracy rate of inference will be higher.

V. Experimental Methods

To thoroughly evaluate the performance of our proposed multi-level knowledge graph augmentation model framework, we conducted additional testing using a breast ultrasound imaging dataset. The dataset was obtained from Sun Yat-sen University Cancer Center [28] and was divided based on the BI-RADS [29] feature criteria. It consists of a total of 1488 ultrasound images of breast cancer tumors. The patients included in the dataset ranged in age from 18 to 73, with an average age of 46. All 1488 breast tumor ultrasound images have been medically verified. Among these images, there are 401 benign tumors, accounting for 26.94% of the total samples, and 1087 malignant tumors, accounting for 73.05% of the total samples. During the data collection process, three ultrasound breast tumor experts from Sun Yat-sen University Cancer Center were invited to conduct BI-RADS scoring to ensure the reliability of the scores. The final scoring results were determined through negotiation when all three votes were the same. To test the generalizability of our model framework, we further conducted testing on two different datasets. The TI-RADS dataset, similar to the BI-RADS dataset, also originates from the Sun Yat-sen University Cancer Center and consists of 2041 samples with six features. Among these samples, there are 1229 positive samples and 1172 negative samples. The POP (post-operative-patient) dataset is a publicly available dataset obtained from the UCI website, containing 90 sets of samples with eight features. By conducting experiments on these diverse datasets, we aim to assess the effectiveness and applicability of our model framework in different scenarios.

*A. Parameter setup*

In order to study the superior performance of the augmented model and further explore the performance of the augmented model under different training data, we did two kinds of experiments. One experiment was to use the TuckER model as the model for the inference part. Make inferences to form your own comparative experiments. The second method we use is to divide the training data sets of different proportions to reflect the inference performance of the augmented model under different data volumes, and set up its own control experiments under each proportion to highlight the effect of the augmented model on the improvement of inference accuracy.

In the above two experiments, the embedding dimension of the entity and relationship of the TuckER model is set to 200, and the model adopts Adam optimizer [30] for parameter optimization, in which the epoch number, learning rate, and input_dropout, hidden_dropout1, hidden_dropout2 rate are set to 200, respectively, 0.0005, 0.3, 0.4, 0.5. The experimental results are shown in TABLE1 and TABLE2.

*B. Self-experimental Comparison*

In order to verify the effect of our proposed multi-level knowledge graph augmentation model framework, we set up our own comparative experiments to verify our superior performance through experiments and theory (Fig.3). The ROC [31] curve is a basic metric widely used for diagnostic prediction evaluation, so in this study, we adopted the ROC curve to evaluate the performance of the KGDA model and the baseline method. The ROC curve plots the sensitivity versus specificity of the predicted outcome for all possible thresholds. Sensitivity here represents the proportion of samples that are correctly predicted among all malignant samples in the test data, while specificity reflects the proportion of samples that are correctly predicted among all benign samples in the test data. ROC curves can also graphically illustrate all the comprehensive trade-offs between sensitivity and specificity at all thresholds. Since the upper-left area of the graph represents the point with high sensitivity and specificity, the closer the curve is to the upper-left corner, the better the performance of the corresponding method. Therefore, we can also use the area under the curve (AUC) value to comprehensively measure the performance, because the closer the curve is to the upper left corner, the larger the AUC value will be. The proportion of samples that are correctly predicted, while the specificity reflects the proportion of samples that are correctly predicted among all benign samples in the test data. The ROC curve can also graphically illustrate all the comprehensive trade-offs between sensitivity and specificity at all thresholds. Since the upper left area of the graph represents the point with high sensitivity and specificity, the closer the curve is to the upper left corner, the better the performance of the corresponding method. Therefore, we can also use the area under the curve (AUC) value to measure performance comprehensively, because the closer the curve is to the upper left corner, the larger the AUC value will be.

*C. Different Ratios Comparison*

In order to verify the performance of the augmented model under different training data sets, we set the ratio of training data to 10%, 30%, 50%, 70%, and 90% of the original data, and tested the data under different ratios. Four indicators of F1, SPE, SEN, and ACC are used as evaluation criteria. The indicators are specifically expressed as follows:

$$\text{ACC} = \frac{TP + TN}{TP + FP + TN + FN}$$

$$\text{SEN} = \frac{TP}{TP + FN}$$

$$\text{SPE} = \frac{TN}{TN + FP}$$

$$F1 = 2 \times \frac{TP}{2*TP+FP+FN}$$

F1, SPE, SEN, ACC, the four indicators that have an internal relationship with each other. For the ACC indicator, it represents the correct rate of the entire data inference, SEN represents the correct rate of positive samples being classified correctly, and SPE represents the correct rate of negative samples being classified correctly. F1 is a comprehensive evaluation of SEN and SPE. The four indicators complement each other during evaluation, which fully reflects the superiority of the performance of the augmented model. At the same time, it can be clearly seen from the chart that the data model augmented by the biclustering algorithm has a greater improvement than the unaugmented data model in the case of a few-sample training data set. It can be considered that the knowledge graph augmentation model based on the biclustering algorithm has a significant improvement in model inference in the case of a few-sample training set.

## VI. RESULTS

*A. Self-experimental Comparison Results*

As shown in the Fig.6, the red curve represents the ROC curve of the unprocessed data model, and the green curve represents the ROC curve (ie. augmentation) of the processed data model. As can be seen from Fig.6, the green curve is above the red curve in most cases, and it can be seen that the effect of the augmented data model is improved compared to the original data model. The performance of the model can be reflected not only from the height of the curve but also from the AUC area to compare the performance of the two. The AUC area of the augmented model reaches 0.98, and the AUC model of the original data model is 0.9759. The superiority of the augmented model can also be seen from the AUC area.

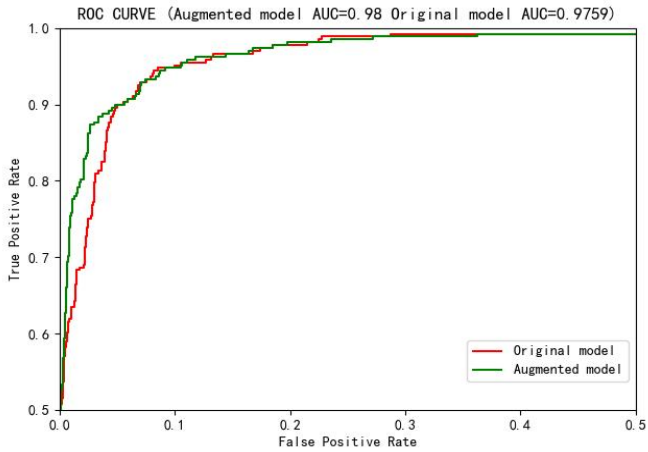

Fig.6. ROC Curve Comparison Chart

*B. Different Ratios Comparison Results*

It can be seen from the Fig.7 that the knowledge graph data model augmented by the biclustering algorithm has a certain degree of improvement in F1, SPE, SEN, and ACC four indicators compared with the knowledge graph data model that has not been augmented by the biclustering algorithm. The Fig.8 is a comparative experiment of the four indicators based on the proportion of training data accounting for 10% of the total data. From the figure, we can see that the inference effect of the data processed by the data augmentation model on the TuckER model is higher than that of the unprocessed data. In the case of the training set, it has a better inference effect.

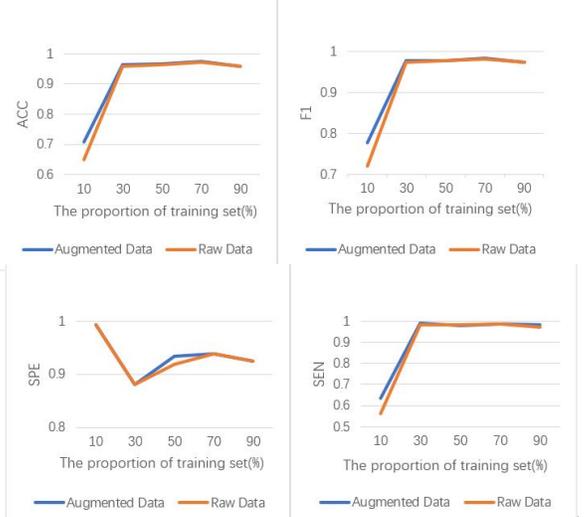

Fig.7. Comparison of different indicators under different data ratios

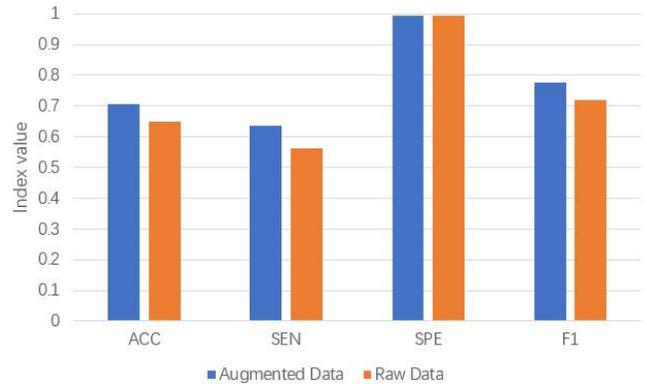

Fig.8.Comparison of various indicators under 10% data ratio. The four methods of ACC, SPE, SEN and F1 are used in the table, and the results prove the advantages of the augmented model in the case of a few samples

The analysis of the evaluation results of the knowledge tensor augmentation method based on biclustering through the ROC curve and the four evaluation indicators of ACC, SEN, SPE and F1 shows that the knowledge tensor augmentation model used in the experiment has a better inference effect on benign and malignant breast cancer than The original data has a certain improvement, especially in the case of a few-sample training set, the knowledge tensor augmentation algorithm based on biclustering has a significant improvement in the knowledge inference effect. It can be seen from the experiment that the inference accuracy rate of the original data is only 64% when the training set only accounts for 10% of the total set, while the data augmented by biclustering, under the same experimental conditions, the inference accuracy rate is 70%. The reason for the significant improvement in the case of a few samples is that because the augmented data is enriched with more and more accurate high-dimensional feature information, in the absence of training samples, it can provide more inference information, so the improvement effect is significant. Later, with the increase of training data, the number of test samples decreases, and the difference between the information rich in the unprocessed data and the processed data decreases, so the improvement effect is less than that in the case of fewer samples, but it is also improved. Through this experiment, it can be proved that the biclustering augmented knowledge tensor augmentation algorithm can significantly improve the inference effect in the case of a few-sample training set.

TABLE 1
Inference accuracy under different training set ratios

|  |  | 10% | 30% | 50% | 70% | 90% |
|---|---|---|---|---|---|---|
| ACC | TuckER | 0.6492 | 0.9587 | 0.9650 | 0.9731 | 0.9597 |
|  | TuckER+KGDA | **0.7074** | **0.9644** | **0.9663** | **0.9753** | 0.9597 |
| SEN | TuckER | 0.5615 | 0.9814 | 0.9817 | 0.9849 | 0.9724 |
|  | TuckER+KGDA | **0.6361** | **0.9888** | 0.9781 | **0.9879** | **0.9816** |
| SPE | TuckER | 0.9926 | 0.8798 | 0.9179 | 0.9385 | 0.9249 |
|  | TuckER+KGDA | 0.9925 | 0.8798 | **0.9333** | 0.9385 | 0.925 |
| F1 | TuckER | 0.7192 | 0.9736 | 0.9764 | 0.9820 | 0.9724 |
|  | TuckER+KGDA | **0.7767** | **0.9773** | **0.9772** | **0.9835** | **0.9727** |

TABLE 2
Inference accuracy of different datasets under 10% training set ratios

|  |  | BI-RADS | TI-RADS | POP |
|---|---|---|---|---|
| ACC | TuckER | 0.6492 | 0.8347 | 0.443 |
|  | TuckER+KGDA | **0.7074** | **0.8662** | **0.5569** |
| SEN | TuckER | 0.5615 | 0.9999 | 0.5 |
|  | TuckER+KGDA | **0.6361** | 0.9999 | **0.5666** |
| SPE | TuckER | 0.9926 | 0.6733 | 0.2631 |
|  | TuckER+KGDA | 0.9925 | **0.7355** | **0.5263** |
| F1 | TuckER | 0.7192 | 0.8566 | 0.5769 |
|  | TuckER+KGDA | **0.7767** | **0.8807** | **0.6601** |

## VII. CONCLUSION

Previous studies have often relied on low-level knowledge for reasoning, which has led to challenges such as limited generalization, poor robustness, and low interpretability. In our work, we propose a multi-level knowledge graph augmentation (KGDA) model framework to address these issues. By extracting pyramidal knowledge from the low-level data, our model overcomes these limitations to a significant extent. Our model leverages the structure of a knowledge graph to enhance the interpretability of reasoning. The key principle of the model framework is to extract pyramidal knowledge from low-level data using a biclustering algorithm. This process improves the overall representation of knowledge, reduces noise interference, uncovers hidden relationships within the low-level data, and enhances the generalization capability of the model. By extracting pyramidal knowledge from the low-level data, our model captures recurring joint representations through the biclustering algorithm. This construction of pyramidal knowledge enhances the interpretability of the model by providing a more structured and understandable representation of the underlying relationships in the data. Additionally, we provide theoretical evidence to support the effectiveness of the model in reducing the variance of knowledge data. This further demonstrates that KGDA can effectively improve the generalization capabilities of the model.Overall, our proposed KGDA model framework tackles the limitations of previous approaches by extracting pyramidal knowledge and enhancing interpretability, thereby improving the generalization and performance of the model.

In this work, we introduce a novel knowledge graph-based data augmentation model called KGDA. Our model leverages the biclustering algorithm to augment the knowledge graph data, which is then combined with the original data using set operations. During the augmentation process, the biclustering algorithm mines low-level features, and multiple high-level features are combined and extracted into a new augmented feature using the Euclidean distance. The augmented data is then fused with the original data using set union operations. To evaluate the advantages of the augmented model, we conducted inference experiments on breast ultrasound data obtained from Sun Yat-sen University Cancer Center. We set up two control groups: one using the augmented data and the other using the original data. Additionally, we performed control experiments using different ratios of data. The experimental results demonstrate that the knowledge graph-based data augmentation model improves the inference performance of the model. Comparative experiments reveal that even with a small amount of data, the knowledge graph-based data augmentation model significantly enhances reasoning effectiveness. This improvement in inference performance highlights the contribution of our work. While our work demonstrates promising results, there is still room for improvement. Future research can expand on the KGDA framework to extract higher-level knowledge and further enhance the model's generalization and interpretability.

ACKNOWLEDGMENT

This work was supported in part by the National Key Research and Development Program under Grant 2018AAA0102100, the National Natural Science Foundation of China under Grants 62071382 and 82030047, and the Innovation Capability Support Program of Shanxi under grant 2021TD-57. Corresponding author: Qinghua Huang.

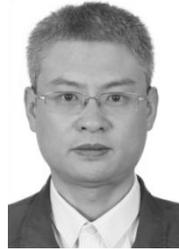

**Qinghua Huang** received the Ph.D. degree in biomedical engineering from the Hong Kong Polytechnic University, Hong Kong, in 2007. Now he is a full professor in School of Artificial Intelligence, Optics and Electronics (iOPEN), Northwestern Polytechnical University, China.

His research interests include multi-dimensional u1trasonic imaging, medical image analysis, machine learning for medical data, and intelligent computation for various applications.

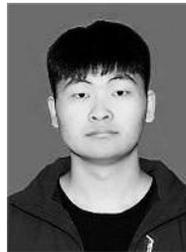

**Yongzhen Wang** is currently pursuing the M.Eng degree in the School of Artificial Intelligence, OPtics and ElectroNics (iOPEN), Northwestern Polytechnical University.

His current research interests include knowledge graph, artificial intelligence.